\documentclass[runningheads]{llncs}
\usepackage[T1]{fontenc}
\usepackage{cite}
\usepackage{amsmath,amssymb,amsfonts}
\usepackage{algorithm}
\usepackage{algpseudocode}
\usepackage{graphicx}
\usepackage{textcomp}
\usepackage{xcolor}
\usepackage{subcaption}
\usepackage[most]{tcolorbox} 
\usepackage{ragged2e}
\usepackage{lipsum}
\usepackage{multirow}
\usepackage{alltt}
\usepackage{enumitem}
\usepackage[font=scriptsize]{caption}
\usepackage{hyperref}
\hypersetup{
    colorlinks=true,
    urlcolor=blue
}
\usepackage{tikz}
\usetikzlibrary{calc}

\begin{document}

\newtcolorbox{llmbox_}[1][]{
    enhanced,
    boxrule=0.6pt,       
    arc=4pt,
    colback=yellow!10,   
    colframe=gray!60,    
    colbacktitle=blue!10!black,
    coltitle=white,
    boxsep=2pt, 
    left=4pt, right=4pt, 
    top=4pt, bottom=4pt,
    toptitle=6pt,
    fontupper=\ttfamily\scriptsize, 
    fonttitle=\ttfamily\bfseries\strut,
    #1                   
}

\title{Agentic Chunking and Bayesian De-chunking of AI Generated Fuzzy Cognitive Maps: A Model of the Thucydides Trap}
\titlerunning{Agentic Chunking and Bayesian De-chunking}
%
\author{Akash Kumar Panda \inst{1} \and
Olaoluwa Adigun \inst{2} \and
Bart Kosko \inst{3}}
\authorrunning{P. Akash et al.}
%
\institute{University of Southern California, Los Angeles, CA 90007, USA  \\ \email{akashpan@usc.edu} \and
Florida International University, Miami, FL 33199, USA \\
\email{olaadigu@fiu.edu}  \and
University of Southern California, Los Angeles, CA 90007, USA \\ \email{kosko@usc.edu}}
\maketitle              

%
\begin{abstract}
\vspace{-0.1in}
We automatically generate feedback causal fuzzy cognitive maps (FCMs) from text by teaching large-language-model agents to break the text into overlapping chunks of text.
Convex mixing of these chunk FCMs gives a representative cyclic FCM knowledge graph.
The text chunks can have different levels of overlap.
The chunk FCMs still mix to form a new FCM causal knowledge graph.
The mixing technique scales because it uses light computation with sparse causal chunk matrices.
The mixing structure allows an operator-level type of Bayesian inference that produces ``de-chunked'' or posterior-like FCMs from the mixed FCM.
These de-chunked FCMs are useful in their own right and allow further iterations of Bayesian updating.
We demonstrate these mixing techniques on the essay text of Allison's ``Thucydides Trap'' model of conflict between a dominant power such as the United States and a rising power such as China. 
The FCM dynamical systems predict outcomes as they equilibrate to fixed-point or limit-cycle attractors.
Seven out of 8 FCM knowledge graphs predicted a type of war when we stimulated them by turning on and keeping on the concept node that stands for the rising power's ambition and entitlement. 
Gemini 3.1 LLMs served as the chunking AI agents.


\end{abstract}

\section{AI Chunking and Bayesian De-chunking of FCMs}

\begin{figure}[!ht]
\centering
\includegraphics[width=0.9\textwidth]{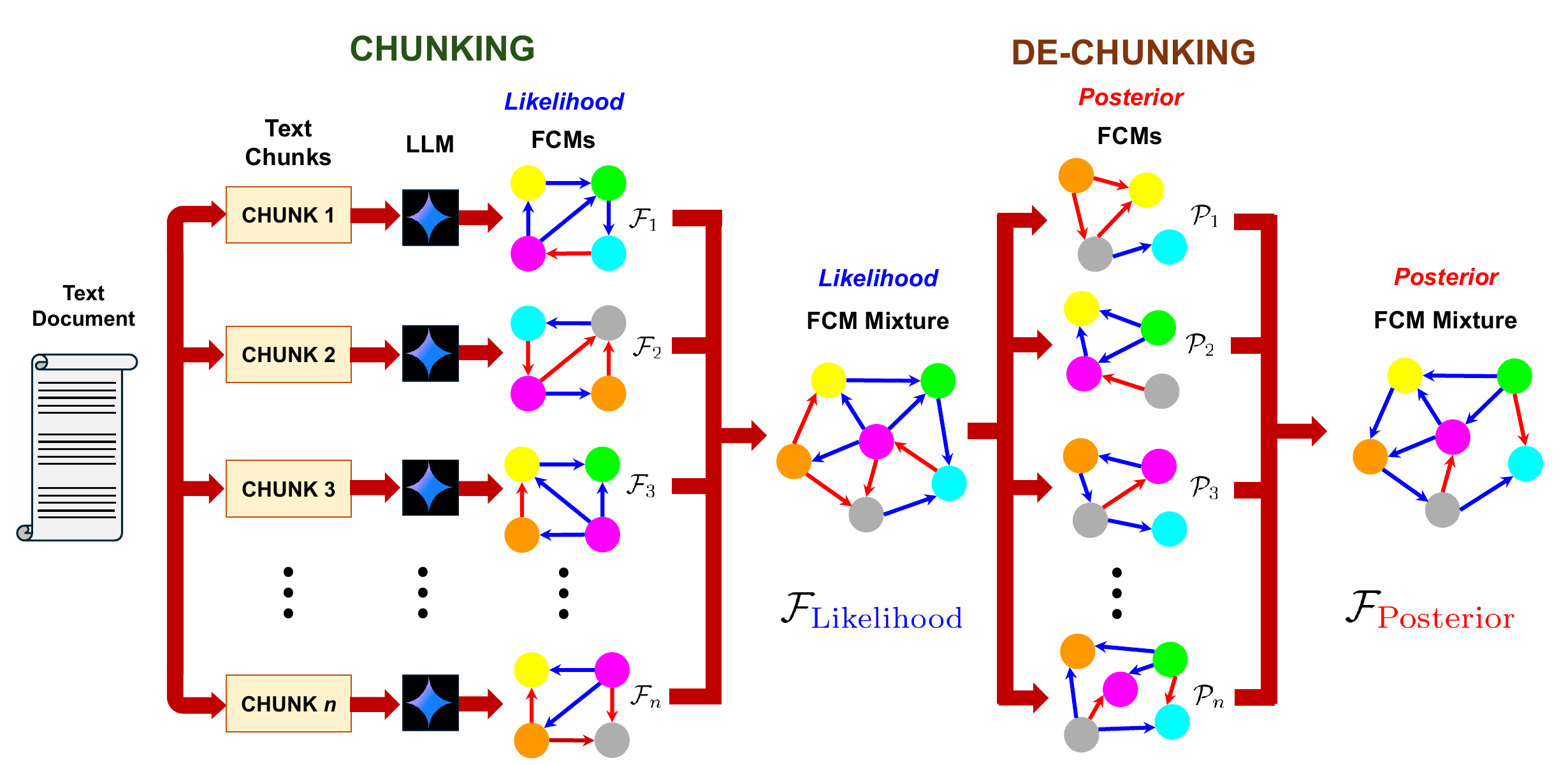}
\caption{\footnotesize{AI-automated chunking and de-chunking of sampled text to create chunk-based Fuzzy Cognitive Maps (FCMs) and the Bayesian-like posterior FCMs. 
The chunking experiment uses Graham Allison's article in \emph{The Atlantic} titled \emph{``The Thucydides Trap: Are the U.S. and China headed for war?''} as input. 
Allison explains:  \emph{``When a rising power is threatening to displace a ruling power, standard crises that would otherwise be contained, like the assassination of an archduke in 1914, can initiate a cascade of reactions that, in turn, produce outcomes none of the parties would otherwise have chosen.''}
The chunking process splits Allison's article into $n$ overlapping chunks of contiguous text. 
Then the large language model (LLM) Gemini 3.1 extracts the respective FCMs $\mathcal{F}_1$, $\mathcal{F}_2$, ... , $\mathcal{F}_n$ from these text chunks using the customized system instructions in Section \ref{sec:map_chunk_to_fcms}. 
These $n$ FCMs then mix through convex combination to give the mixed FCM $\mathcal{F}_{\text{Likelihood}}$. 
The de-chunking process finds the posterior FCM $\mathcal{P}_k$ that corresponds to the likelihood FCM $\mathcal{F}_k$. 
Then the posterior FCMs also mix through convex combination to give $\mathcal{F}_{\text{Posterior}}$.}  \vspace{-0.3in}}
\label{fig:Chunking-and-De-chunking}
\end{figure}

We show how to use large-language-model agents to break sampled text into overlapping chunks and then map those text chunks to sparse feedback fuzzy cognitive maps (FCMs).
The technique offers a practical way to decompose large text documents into representative knowledge graphs and then recombine them into a unified graph that represents the entire document.
Users can then ask this combined FCM network what-if causal questions and get equilibrium answers.

The FCMs are fuzzy because their local directed causal edges take values in the bipolar interval $[-1, 1]$ and represent partial causality or degrees of causal decrease or increase.
The FCMs are cyclic knowledge graphs that define nonlinear feedback dynamical systems whose equilibria give fixed-point or limit-cycle answers to the what-if questions that users impose on their causal concept nodes.

\begin{figure}[t]
\centering
\includegraphics[width=0.95\textwidth]{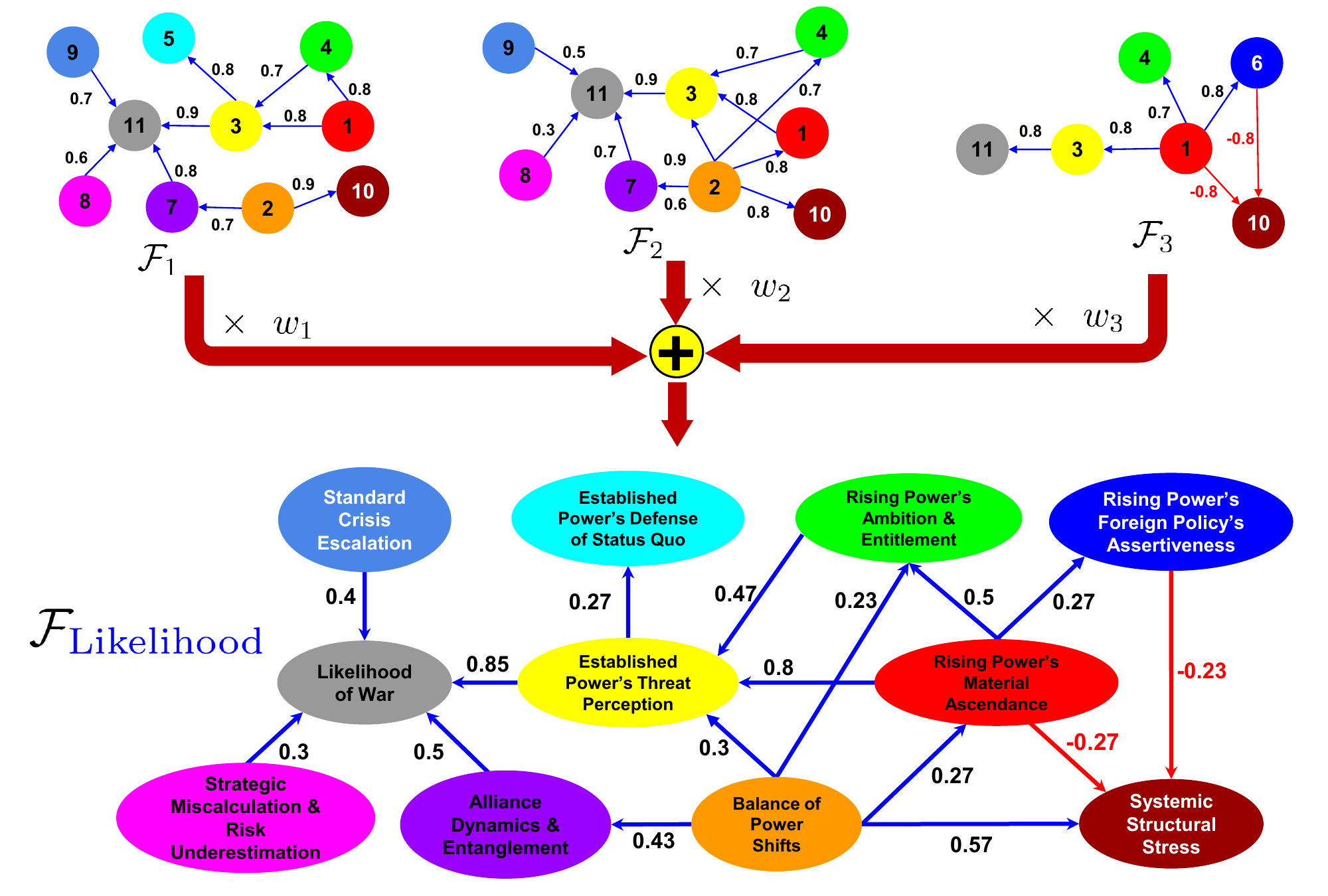}
\caption{\footnotesize{Likelihood FCM Mixture. 
The 3 likelihood FCMs $\mathcal{F}_1$, $\mathcal{F}_2$, and $\mathcal{F}_3$ extracted from chunks of the ``Thucydides Trap'' article are on the top. 
The 11-node mixture FCM $\mathcal{F}_{\text{Likelihood}}$ mixes the 3 likelihood FCMs with equal mixing weights $\frac{1}{3}$.} \vspace{-0.20in}}
\label{fig:Mixed-likelihood-FCMs-v1}
\end{figure}

Figure  \ref{fig:Chunking-and-De-chunking} shows how text from Allison's ``Thucydides Trap'' essay \cite{allison2015thucydides} on Great-
Power conflict leads to 3 chunk FCMs that combine into a representative FCM for the entire sampled document.
Figure \ref{fig:Mixed-likelihood-FCMs-v1}  shows how these chunk FCMs combine to form a representative FCM knowledge graph by mixing their zero-padded edge matrices with convex mixing weights or probabilities.
This automated and mixed Thucydides FCM differs in kind from the earlier hand-drawn FCM in \cite{osoba2019beyond}.

\begin{figure}[!ht]
\centering
\includegraphics[width=0.9\textwidth]{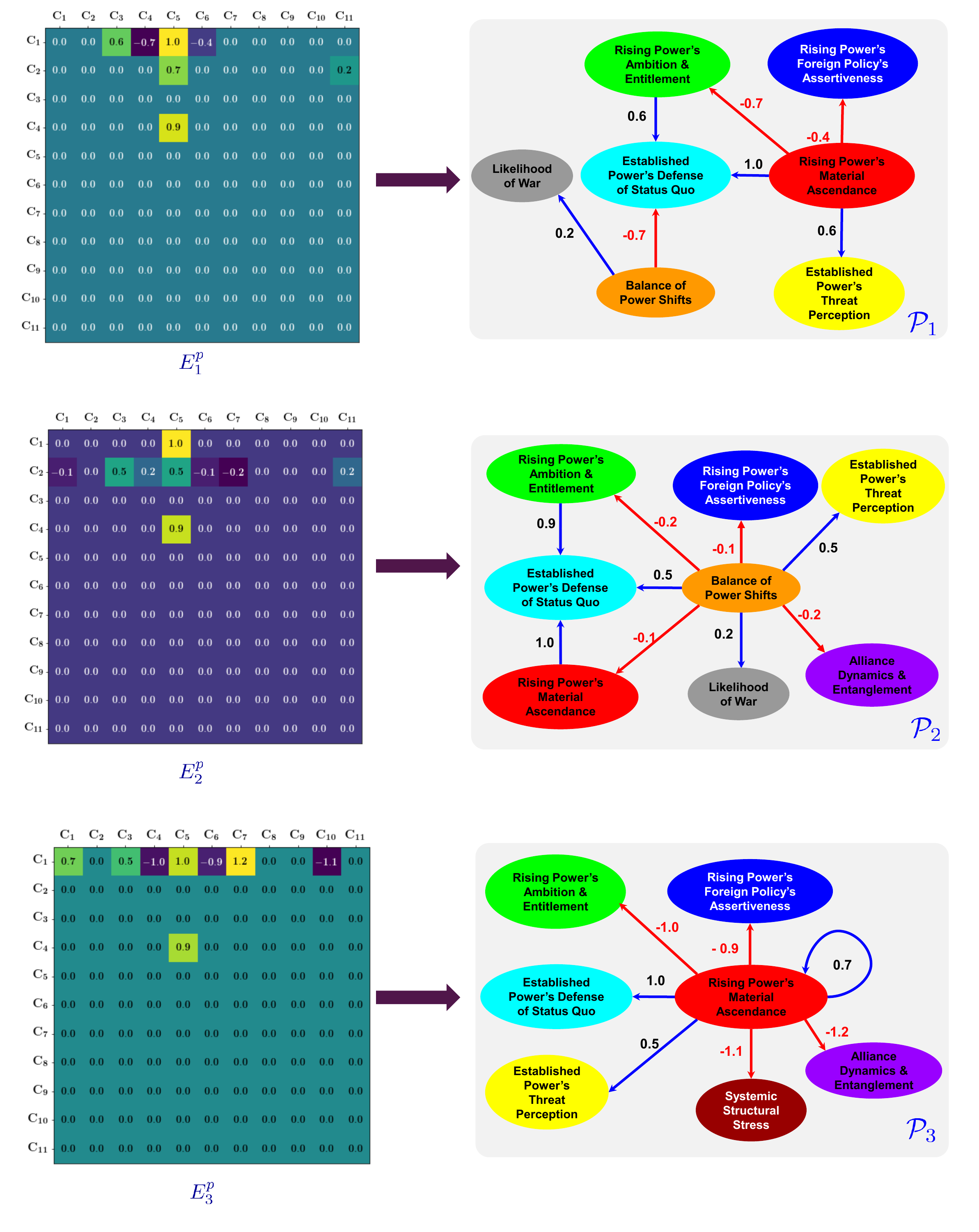}
\caption{\footnotesize{FCM De-chunking and Bayesian-like posterior FCMs.  Equation (\ref{eqn:4}) converts the mixed likelihood FCM edge matrix $E^l$ into 3 sparse posterior-like edge matrices $E^p_1$, $E^p_2$, and $E^p_3$.  These posterior matrices then define the posterior FCMs $\mathcal{P}_1$, $\mathcal{P}_2$, and $\mathcal{P}_3$.  \vspace{-0.0in}}}
\label{fig:Mixed-posterior-FCMs}
\end{figure}

We further show that we can de-mix or ``de-chunk'' this mixed FCM to give Bayesian-like (operator-based) posterior FCMs as in Figure \ref{fig:Mixed-posterior-FCMs}.
These FCMs can then participate in future bouts of Bayesian-like updating and mixing.

The next sections explain how FCM dynamical systems work and how to guide LLM agents to perform semantic-split-like text chunking and the crucial mapping from text chunks to combinable FCMs.  
We demonstrate these techniques on Allison's popular essay that addresses whether the United States and China may get caught in a warlike Thucydides Trap.
\vspace{-0.0in}

\section{Fuzzy Cognitive Maps}
\label{sec:fcms}

FCMs model causal dynamical systems as directed weighted graphs \cite{kosko1986differential, kosko1986fuzzy, kosko1988hidden, osoba2017fuzzy, ziv2018potential, glykas2010fuzzy, papageorgiou2013fuzzy, stach2010divide ,taber2007quantization, panda2025causal, panda2025agentic}. 
The concept nodes describe the causal variables in the dynamical system and the directed edges describe the causal relationships between those nodes. 
FCMs allow feedback and therefore converge to non-trivial equilibria like limit cycles. 
FCMs model dynamical systems by approximating their underlying maps from inputs to equilibria.

In contrast:  Bayesian belief networks do not mix in general.  
They are directed-\emph{acyclic} graphs that lack feedback and so have no dynamics.  
Inference requires marginalization of joint probabilities that involves NP-hard computation \cite{cooper1990computational}.

\subsection{Causal Edge Matrix}

The directed edges of the FCM describe the causal relationships between concept nodes. 
An edge $e_{ij}$ from the $i^\text{th}$ concept node $C_i$ to the $j^\text{th}$ concept node $C_j$ means ``$C_i$ causes $C_j$''. 
The fuzzy edge-weights describe partial causality. 
The edge weight $e_{ij} \in [-1,1]$ on the edge from the $i^\text{th}$ node to the $j^\text{th}$ node gives the degree to which $C_i$ causes $C_j$:  $e_{ij} = Degree(C_i \rightarrow C_j).$

A positive $e_{ij}$ means that $C_j$ increases when $C_i$ increases and a negative $e_{ij}$ means that $C_j$ decreases if $C_i$ increases. 
The weight $e_{ij}$ is zero when there is no causal edge between $C_i$ and $C_j$. 
The magnitude of $e_{ij}$ is high if there is a strong causal relationship between $C_i$ and $C_j$. 
Low magnitude of $e_{ij}$ describes a weak causal relationship between $C_i$ and $C_j$.

An $n \times n$ matrix $E$ describes all the directed weighted edges of a $n$-node FCM. 
The edge weight $e_{ij}$ corresponds to the matrix element on the $i^\text{th}$ row and the $j^\text{th}$ column. 
The matrix element is zero if there is no edge between the corresponding node-pair. 

\subsection{FCM Evolution}

A $n$-dimensional row vector $C(t) \in [0,1]^n$ describes the state of the FCM's concept nodes at time $t$. 
The $i^\text{th}$ node is ``active'' or ``on'' at time $t$ if the $i^\text{th}$ component $C_i(t)$ of the state vector $C(t)$ is equal to or close to one. 
The $i^\text{th}$ node is ``inactive'' or ``off'' at time $t$ if $C_i(t)$ is equal to or close to zero. 
A node is partially active otherwise. 
The causal variables corresponding to the active nodes are present in the system and those corresponding to the inactive nodes are absent. 
The causal factors are partially present in the system if their corresponding nodes are partially active. 

FCMs evolve in discrete time through vector-matrix multiplication and nonlinear squashing. 
The state $C_j(t+1)$ of the $j^\text{th}$ concept node $C_j$ at discrete time step $t+1$ is:
\vspace{-0.1in}
\begin{align}\label{eq:fcm-update}
    C_j(t+1) = \Phi\bigg(\sum_{i=1}^n  C_i(t)e_{ij}\bigg)
\end{align}
where $\Phi$ is a nonlinear function bounded between zero and one. 

The sum $\sum_{i=1}^n  C_i(t) e_{ij}$ is the matrix product between the state row-vector $C(t)$ and the edge matrix $E$. 
The nonlinear function $\Phi$ then squashes this product between zero and one. 

This process repeats itself to give the discrete-time evolution of the FCM. 
The FCM starts with the initial state $C(0)$ at time $t = 0$ and then goes through the states $C(1)$, $C(2)$, $C(3)$, and so on in order. 
The active nodes in this state-vector sequence qualitatively describe the trajectory of the dynamical system that the FCM models. 

\subsection{FCM Equilibria}

The equilibria characterize a dynamical system. 
The equilibrium behavior of the FCM depends on the limiting behavior of the state-vector sequence. 
The FCM converges to a ``fixed point'' if the state-vector sequence converges to a constant vector. 
The FCM converges to a $K$-step ``limit cycle'' for an integer $K > 1$ if $C(t+K) = C(t)$ somewhere in the state-vector sequence. 
Then the FCM converges to an equilibrium where $K$ state vectors repeat themselves over and over in the same order. 
The FCM may also converge to a chaotic attractor where there are no repeating patterns in the state-vector sequence. 

The set of all initial conditions $C(0)$ that lead to a given equilibrium describes the ``basin of attraction'' for that equilibrium. 
The FCM describes a map from these basins to their corresponding equilibrium attractors. 
The basins of the FCM's equilibria partition the FCM's input space. 
The FCM models a dynamical system by approximating its corresponding basin-to-equilibrium map. 

\subsection{Clamping and Pulsing}

``Clamping'' the $k^\text{th}$ node $C_k$ to a value $c_k \in [0,1]$ forces $C_k(t)$ to equal $c_k$ irrespective of the values of other nodes at time $t-1$. 
The $k^\text{th}$ node is clamped ``on'' if $c_k \approx 1$ and it is clamped ``off'' if $c_k \approx 0$. 
Clamping acts as a forcing function and pushes the FCM into new equilibria. 
FCMs can clamp one or more nodes at a time. 
Clamping is often a way to implement policies and answer What-if questions regarding their effects. 

FCMs can also `pulse' a node instead of clamping it. 
`Pulsing' the $k^\text{th}$ node $C_k$ to a value $c_k$ at time $t_0$ forces $C_k(t_0)$ to equal $c_k$ irrespective of the values of other nodes at time $t_0-1$. 
This only affects the FCM at time $t_0$ and the FCM reverts back to its unperturbed dynamics after $t_0$. 
The node $C_k$ pulses `on' at time $t_0$ if $C_k(t_0) \approx 1$ and it pulses `off' at time $t_0$ if $C_k(t_0) \approx 0$. 
FCMs can also pulse more than one node at a time. 

\subsection{FCM Mixtures}

FCMs combine their knowledge through convex mixing. 
Consider $m$ FCMs. 
Say the $k^\text{th}$ FCM has the set of nodes $S_k$ and the edge matrix $E_k$. 
The node-set $S$ for the $N$-node FCM mixture is $S_1\cup S_2\cup S_3\cup ...\cup S_m$. 
The $N\times N$ matrix $\Tilde{E_k}$ pads the $k^\text{th}$ edge matrix $E_k$ with zero rows and zero columns corresponding to the nodes in the set difference $S - S_k$. 
The edge matrix $E$ for the mixed FCM is
\begin{align}
    E = \sum_{k=1}^m w_k\Tilde{E_k} \label{eg:fcm_mixing}
\end{align}
where $w_k$ are convex mixing weights such that $w_k\geq0$ and $\sum_{k=1}^mw_k=1$. 
FCM mixing is closed:  Mixing FCMs always gives back an FCM.

\subsection{Posterior FCMs from the Ratio Structure of Bayes Theorem}

\emph{Total probability} mixes likelihoods $P(E|H_j)$ using the prior probabilities $P(H_j)$ as convex mixing weights:  $P(E) = \sum_{j=1}^m P(H_j)P(E|H_j)$ for evidence $E$ and disjoint exhaustive hypotheses $H_j$. 
Then Bayes theorem gives the  posterior probabilities $P(H_k|E)$:  $P(H_k|E)=P(H_k)P(E|H_k)/P(E)$ that further simplifies to  $P(H_k|E)=P(H_k)P(E|H_k)/\sum_{j=1}^m P(H_j)P(E|H_j)$ because of total probability. 
The Bayes posterior $P(H_k|E)$ has the ratio form $\frac{a}{a+b} = a (a + b)^{-1}$ and this we can generalize with FCM nonlinear operators and their causal edge matrices.

Equation (\ref{eg:fcm_mixing}) shows that mixed FCMs also share the same convex mixing structure as does total probability. 
But they mix edge \emph{matrices} instead of mixing likelihood probabilities.
This gives a new Bayesian-like posterior FCM. 
Say $m$ likelihood FCMs $F_1$, $F_2$, ... , $F_m$ with edge matrices $E^l_1$, $E^l_2$, ... , $E^l_m$ mix with respective mixing weights $w_1$, $w_2$, ... , $w_m$. 
Then the posterior FCM $P_k$ that corresponds to the $k^\text{th}$ FCM $F_k$ has the edge matrix $E^p_k$: 
\begin{align}\label{eqn:4}
    E^p_k &= w_k\tilde{E}^l_k(E^l)^{-1} = w_k\tilde{E}^l_k\Bigg(\sum_{j=1}^mw_j\tilde{E}^l_j\Bigg)^{-1}
\end{align}
if the mixed-FCM's edge matrix $E^l=\sum_{j=1}^mw_j\tilde{E}^l_j$ is invertible.
So the $m$ posterior edge matrices $E^p_k$ sum to the identity matrix $I$ just as the $m$ posteriors $P(H_k|E)$ sum to $1$:   $\sum_{k=1}^m E^p_k  = \bf{I}$.
The posterior FCMs also mix to give the posterior-FCM mixture $\mathcal{F}_{\text{Likelihood}}$ that gives an operator form of total probability:  $E^P=\sum_{k=1}^m v_kE^p_k$ for convex mixing weights: $v_k \geq 0$ and $\sum_{k=1}^mv_k=1$. 
 
If $E^l$ is not invertible then we use the \emph{pseudo-inverse} of $E^l$:
\begin{align}
    E^p_k &= w_k\tilde{E}^l_k(E^l)^T(E^l(E^l)^T)^{-1} \\
    &= w_k\tilde{E}^l_k\Bigg(\sum_{j=1}^mw_j\tilde{E}^l_j\Bigg)^T\Bigg(\Bigg(\sum_{j=1}^mw_j\tilde{E}^l_j\Bigg)\Bigg(\sum_{j=1}^mw_j\tilde{E}^l_j\Bigg)^T\Bigg)^{-1} \label{eq:posterior_FCM}
\end{align}
The posterior FCM may have a few edge weights outside the bipolar interval $[-1,1]$. 
We can set their magnitude equal to 1 or renormalize.

\section{Mapping Text Chunks to FCMs}
\label{sec:map_chunk_to_fcms}

This section presents the multi-step process that converts text to FCMs. 
The process breaks a full text into small chunks and uses an LLM agent to extract the nodes and fuzzy edge weights for each corresponding FCM.

\vspace{-1em}
\begin{figure}[!ht]
\centering
\includegraphics[width=0.80\textwidth]{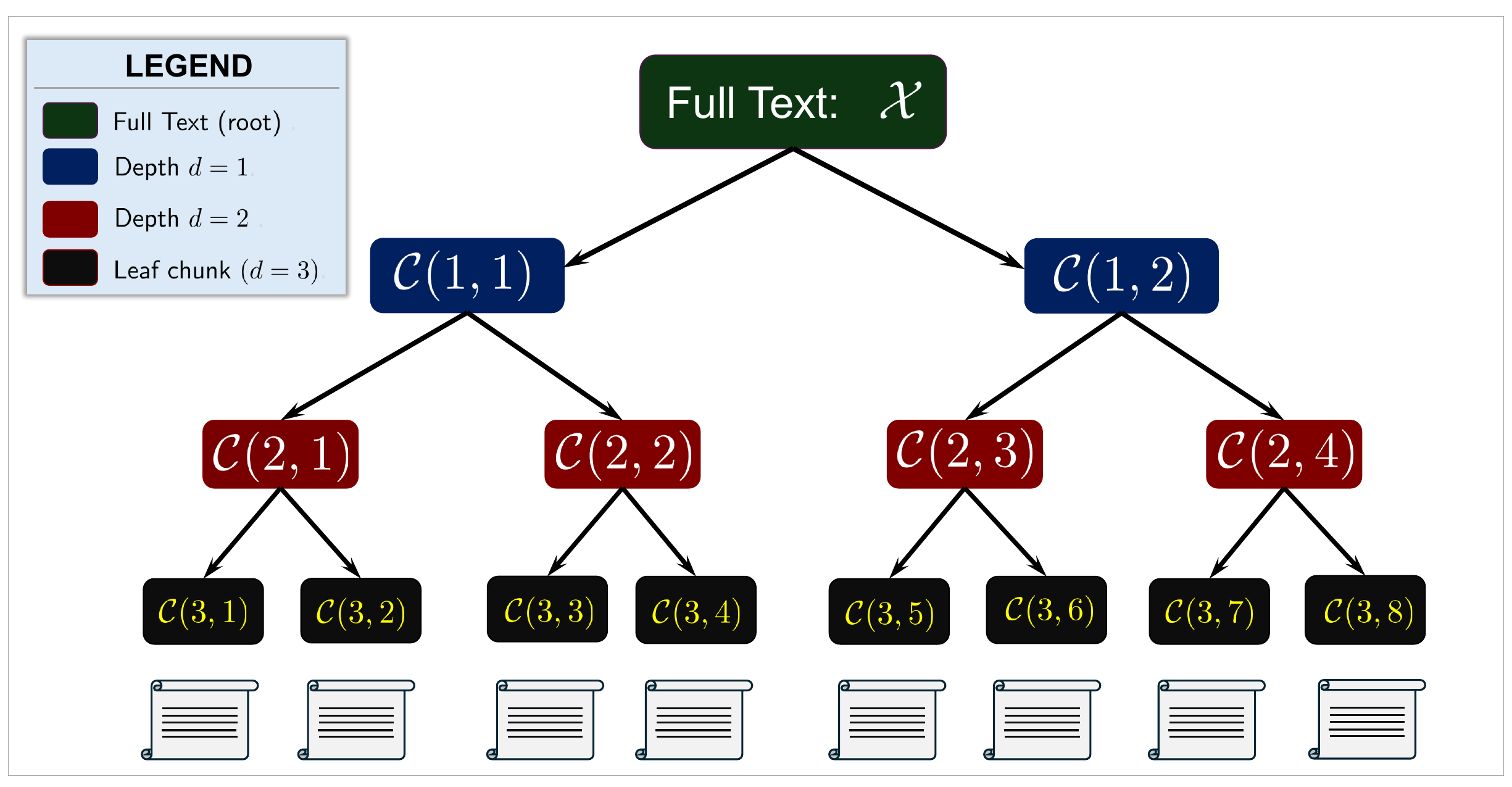}
\caption{\footnotesize{Binary recursive text splitting: The  binary recursive text method breaks down an input text $\mathcal{X}$ to small, contiguous, and non-overlapping text chunks. 
This method uses a binary tree structure and the text chunks are the leaf nodes at depth $d_{\max}$.
The maximum depth $d_{\max} = 3$ so the number of text chunks is $2^3 = 8$.
}\vspace{-0.4in}}
\label{fig:binary_recursive_splitting}
\end{figure}

\subsection{Text Chunking}
Text chunking is the process of breaking a text down into small, non-overlapping or overlapping text chunks.
We define two types of text chunking: non-overlapping chunking and overlapping chunking.
Both techniques use a simple binary recursive text splitting method to generate contiguous and small text chunks. 
Non-overlapping chunking does not generate any overlapping chunks while overlapping chunking does. \\

\noindent {\bf{Binary Recursive Text Splitting:}}
This method uses a binary tree to recursively divide a text based on the number of paragraphs in small text chunks.
The division continues until it reaches the given user-specified maximum depth $d_{\max}$.
The leaves of the binary tree are the small and non-overlapping text chunks of the input text.
The union or concatenation of these chunks gives back the text at any depth of the binary tree.

The binary text splitting uses the $\textsc{Split}$ function to recursively break down a given input text $\mathcal{X}$ with $N$ paragraphs and a maximum depth $d_{\max}$.
The range of $\mathcal{X}$ is $[\mathtt{lo},\mathtt{hi}]$  with $\mathtt{lo} =1 $ and $\mathtt{hi} = N$. 
This is the initial range for the splitting and the corresponding depth is $d = 0$.   
The $\textsc{Split}$ function bisects the current range at the midpoint: $\mathtt{mid }= \lfloor (\mathtt{lo} + \mathtt{hi})/2 \rfloor$. 
The method then applies $\textsc{Split}$ to these two ranges: $[\mathtt{lo}, \mathtt{mid}]$ and $[\mathtt{mid}+1, \mathtt{hi}]$, and updates the $d$.
Each node of receives the index $\mathcal{C}(d,k)$ where $d$ is the depth and $k$ is its left-to-right position.
This produces $2^{d_{\max}}$ non-overlapping chunks $\mathcal{C}(d_{\max}, 1) , ..., \mathcal{C}(d_{\max}, 2^{d_{\max}})$ that partitions $\mathcal{X}$.
Figure ~\ref{fig:binary_recursive_splitting} shows an example of how this splitting method works. \\

\noindent {\bf{Non-overlapping Text Chunking:}}
The non-overlapping chunking uses the binary recursive splitting method to divide $\mathcal{X}$ into $2^{d_{\max}}$ small and non-overlapping chunks.
Algorithm ~\ref{alg:nonoverlapping_chunking} shows the pseudo-code for this chunking method.
It runs recursively and returns the non-overlapping chunks or their corresponding paragraph ranges at the leaf nodes.
\vspace{-0.2in}

\begin{algorithm}
\caption{Non-overlapping Chunking}
\label{alg:nonoverlapping_chunking}
\begin{algorithmic}[1]
\Require Text $\mathcal{X}$ with $\mathtt{N}$ paragraphs, maximum depth $d_{\max}$
\Ensure  Set of non-overlapping chunks
         $\{\mathcal{C}(d_{\max}, k) : k = 1, \dots, 2^{d_{\max}}\}$
 \Statex
\State \Call{Split}{$\mathtt{1},\ \mathtt{N},\ 1,\ 1$}  \Comment{\textcolor{blue}{Recursive binary splitting}}
\Statex
\Function{Split}{$\mathtt{lo},\ \mathtt{hi},\ d,\ k$}
    \If{$d = d_{\max}$}
        \State $\mathcal{C}(d_{\max},\, k) \leftarrow \mathcal{X}[\mathtt{lo} \,{:}\, \mathtt{hi}]$
        \State \Return
    \EndIf
    \State $\mathtt{mid} \leftarrow \lfloor (\mathtt{lo} + \mathtt{hi}) / 2 \rfloor$
    \State \Call{Split}{$\mathtt{lo},\ \mathtt{mid},\ d+1,\ 2k-1$}
    \State \Call{Split}{$\mathtt{mid + 1},\ \mathtt{hi},\ d+1,\ 2k$}
\EndFunction
\end{algorithmic}
\end{algorithm}
\vspace{-0.2in}



The size $N(d, k)$ of  chunk $\mathcal{C}(d,k)$ is the number of paragraphs in the chunk.
The text chunks at any depth $d$ are such that:
\vspace{-0.15in}
\begin{align}
 \sum_{k=1}^{2^d} \big{|} \mathcal{C}(d, k) \big{|} &=   \sum_{k=1}^{2^d} N(d, k) = N
\end{align}
where $k,l \in \{1,2,...,2^d\}$. \\

\noindent {\bf{Overlapping Text Chunking:}}
This method breaks down an input text to overlapping text chunks.
These are the inputs for this method: input text $\mathcal{X}$, maximum depth $d_{\max}$, and split factors $\alpha_1$ and $\alpha_2$. 
The method returns $2 \cdot {2^{d_{\max} } }- 1$ text chunks in total.
Algorithm \ref{alg:full-chunking} shows the pseudo-code for this method.

This chunk methods involves two steps. 
The first step uses the binary recursive text splitting method to break $\mathcal{X}$ to $2^{d_{\max}}$ non-overlapping chunks. 
These are the leaf nodes for the corresponding binary tree. 
The second step then creates $2^{d_{\max}} - 1$ overlapping chunks from the non-overlapping chunks.
The degree of overlap for the overlapping chunks depends on the values of $\alpha_1$ and $\alpha_2$.
\vspace{-0.20in}

\begin{figure}[!ht]
\centering
\includegraphics[width=0.85\textwidth]{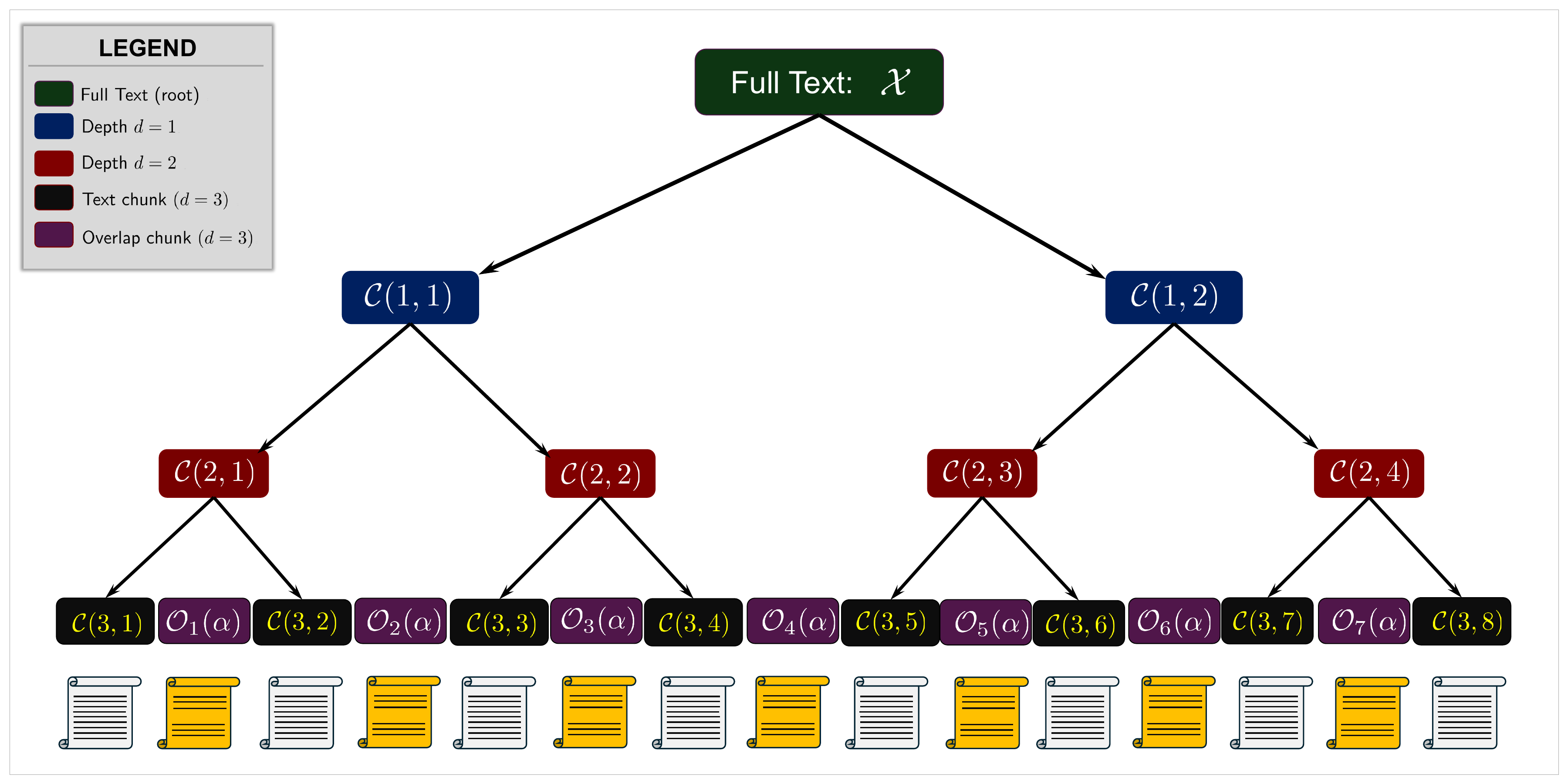}
\caption{\footnotesize{Overlapping text chunking: 
This illustrates the overlapping chunking method (Algorithm \ref{alg:full-chunking}) with $d_{\max} = 3$. 
This breaks down an input text $\mathcal{X}$ to a sequence of overlapping chunks. 
The first step applies the binary recursive text split to generate the non-overlapping chunks: $\mathcal{C}(3,1),...,\mathcal{C}(3,8)$. 
The second step extracts proportional overlapping chunks $\mathcal{O}_k(\alpha)$ for the non-overlapping chunks.
}}
\label{fig:overlap_chunking}
\end{figure}
\vspace{-0.2in}

 Each text chunk breaks down to two disjoint sub-chunks: a leading sub-chunk and a trailing sub-chunk.
The size of this sub-chunks depend on the split factor $\alpha$.
The leading sub-chunk $ \mathcal{L}(\mathcal{C}(d, k), \alpha)$ of $\mathcal{C}(d, k)$ with split factor $\alpha$ is the first $\lceil \alpha \cdot  N(d,k) \rceil$ paragraphs of $\mathcal{C}(d, k)$.
We have $\ell(d,k) =  \lceil \alpha \cdot  N(d,k) \rceil $ and $\mathcal{L}(\mathcal{C}(d, k), \alpha)  = \mathcal{C}(d, k)[1: \ell(d,k)] $. 

The trailing sub-chunk $ \mathcal{T}(\mathcal{C}(d, k), \alpha)$ of $\mathcal{C}(d, k)$ with the same split factor  is the last $\lfloor (1-\alpha) \cdot  N(d,k) \rfloor$ paragraphs of $\mathcal{C}(d, k)$.
We have $\tau(d,k) = \lfloor (1-\alpha) \cdot  N(d,k) \rfloor$ and $\mathcal{T}(\mathcal{C}(d, k), \alpha) = \mathcal{C}(d, k)[\tau(d,k): N(d,k)]$ .

The overlapping text chunking method generates overlapping text chunks from the non-overlapping text chunks.
The first step yields $2^{d_{\max}}$ non-overlapping chunks:  $\mathcal{C}(d_{\max}, 1)$, ...., $\mathcal{C}(d_{\max}, 2^{d_{\max}})$.
Let $\mathcal{O}(k, \alpha_1, \alpha_2)$ represent the $k^{th}$ overlapping chunk .
 $\mathcal{O}_k(\alpha_1, \alpha_2)$ is the union of the trailing sub-chunk of $\mathcal{C}(d_{\max},k)$ at split factor $\alpha_1$  and the leading sub-chunk of $\mathcal{C}(d_{\max},k+1)$ at split factor $\alpha_2$.
 We have $\mathcal{O}_k(\alpha_1, \alpha_2) =  \mathcal{T}(\mathcal{C}(d, k), \alpha_1) \cup \mathcal{L}(\mathcal{C}(d, k+1), \alpha_2) $ where $k \in \{1, 2,..,2^d - 1\}$.

A special case is the proportional overlap chunk $\mathcal{O}_k(\alpha)$.
This is  equivalent to $\mathcal{O}_k(\alpha_1, \alpha_2)$ with $\alpha_1 = 1 - \alpha$ and $\alpha_2 = \alpha$. 
The trailing sub-chunk of  $\mathcal{C}(d, k)$ and the leading sub-chunk of $\mathcal{C}(d, k+1)$ have similar paragraph proportions relative to their respective chunks.
We have $ 
\mathcal{O}_k(\alpha) = \mathcal{O}_k(1-\alpha, \alpha) = \mathcal{T}(\mathcal{C}(d, k), 1-\alpha) \cup \mathcal{L}(\mathcal{C}(d, k+1), \alpha)$ .
\vspace{-0.2in}

\begin{algorithm}[h]
\caption{Overlapping Chunking}
\label{alg:full-chunking}
\begin{algorithmic}[1]
\Require Text $\mathcal{X}$ with $\mathtt{N}$ paragraphs, maximum depth $d_{\max}$,
         split factors $\alpha_1, \alpha_2 \in [0, 1]$
\Ensure  Non-overlapping chunks $\{\mathcal{C}(d_{\max}, k) : k = 1, \dots, 2^{d_{\max}}\}$
         and overlapping chunks $\{O_k(\alpha_1, \alpha_2) : k = 1, \dots, 2^{d_{\max}} - 1\}$

\Statex
\State \Call{Split}{$1,\ \mathtt{N},\ 1,\ 1$}  \Comment{\textcolor{blue}{Step 1: Recursive binary splitting}}
\Statex
\Function{Split}{$\mathtt{lo},\ \mathtt{hi},\ d,\ k$}
    \State $\mathcal{C}(d,\, k) \leftarrow \mathcal{X}[\mathtt{lo} \,{:}\, \mathtt{hi}]$
    \If{$d = d_{\max}$}
        \State \Return
    \EndIf
    \State $\mathtt{mid} \leftarrow \lfloor (\mathtt{lo} + \mathtt{hi}) / 2 \rfloor$
    \State \Call{Split}{$\mathtt{lo},\ \mathtt{mid},\ d+1,\ 2k-1$}
    \State \Call{Split}{$\mathtt{mid}+1,\ \mathtt{hi},\ d+1,\ 2k$}
\EndFunction

\Statex
\For{$k = 1$ \textbf{to} $2^{d_{\max}} - 1$} \Comment{\textcolor{blue}{Step 2: Construct overlapping chunks}}
    \State $N(d_{\max}, k) = |\mathcal{C}(d_{\max},\, k)| $ 
    \State $N(d_{\max}, k+1) = |\mathcal{C}(d_{\max},\, k+1)| $
    \State $t \leftarrow \lfloor (1 - \alpha_1) \cdot N(d_{\max},\, k) \rfloor$
    \State $\ell \leftarrow \lceil \alpha_2 \cdot N(d_{\max},\, k+1) \rceil$
    \State $\text{trail}_k     \leftarrow$ last $t$ paragraphs of $\mathcal{C}(d_{\max},\, k)$
    \State $\text{lead}_{k+1} \leftarrow$ first $\ell$ paragraphs of $\mathcal{C}(d_{\max},\, k+1)$
    \State $\mathcal{O}_k(\alpha_1, \alpha_2) \leftarrow \text{tail}_k \cup \text{lead}_{k+1}$
\EndFor
\end{algorithmic}
\end{algorithm}
\vspace{-1.5em}

\subsection{Local FCM  Extraction}

This focus on extracting an FCM for each text chunk with an LLM agent.
Each  process involves three sequential steps: (1) concept identification, distillation, and normalization, (2) causal validation, and (3) fuzzy edge weighting. 
The system prompt instructs the LLM agent to run these sequential steps.
The agent then returns the corresponding local FCM.\\

\noindent{\bf{Local Node Definition}:}
The step takes in a text chunk and returns a list of nodes for its corresponding FCM.
The process follows three sequential steps: \emph{identification}, \emph{distillation}, and \emph{normalization}. 
Identification step finds every entity or idea that logically increases or decreases within the context of the text chunk. 
Distillation step reduces nouns by removing modifiers and attributive adjectives from all identified entities. 
This step merges synonymous entities into one and resolves pronouns to their respective noun antecedents.\\

\noindent{\bf {Causal Validation}:}
This step takes the list of nodes from the local node definition and filters out dead nodes.
Dead nodes are entities that do not have a causal link to or from another entity.
The causal validation process follows two sequential steps: \emph{causal verification} and \emph{causal discarding}.
 The causal verification step checks every node in the list against the text chunk.
The causal discarding step removes any remaining node that lacks a causal link within the context of their respective text chunk.\\

\noindent{\bf {Fuzzy Edge Weighting}:}
This step takes in a list of causally validated nodes and their causal relationships and returns a weighted edge list for its corresponding FCM. 
The process follows two sequential steps: \emph{direction assignment} and \emph{intensity scoring}. 
The direction assignment step determines the polarity of each relationship.
It assigns a positive weight when the cause and effect move in the same direction and a negative weight when they move in opposite directions. 
The intensity scoring step then assigns a numerical weight based on the linguistic strength of the relationship in the text chunk.\\

\noindent The LLM agent returns the local FCM $\mathcal{F}_{d}(k)$ for input chunk $\mathcal{C}({d},k)$. 
This local extraction runs over chunks $\mathcal{C}(d, 1)$,...., $\mathcal{C}(d, 2^{d})$.
Their respective local FCMs are $\mathcal{F}_{d}(1)$,..., $\mathcal{F}_{d}({2^d})$.
We now  mix these local FCMs into a global FCM. 

\subsection{Node Consolidation}

This process combines the node lists of multiple local FCMs into a single global node list. 
The process follows three sequential steps: (1) global deduplication, (2) thematic clustering, and (3) intra-theme semantic merging. \\

\noindent{\bf {Global Deduplication}:}
The step combines the node lists of all local FCMs into a single global list.
This list may contain duplicate nodes.
The LLM agent then scans the full list and eliminates redundancy at two levels. 
Exact string matches merge directly into a single surviving node. 
Near-duplicate nodes that share the same underlying meaning despite minor lexical differences also merge.\\

\noindent{\bf{Thematic Clustering}:}
The step takes the deduplicated node list and organizes it into high-level themes. 
It analyzes the semantic domain of each node and groups related nodes under a common theme.
Every node must belong to exactly one theme. \\

\noindent{\bf{Intra-theme Semantic Merging}:}
The step examines the nodes within each theme and identifies those that represent different facets of the same underlying causal concept.
It groups these nodes into semantic clusters and merges each cluster into a single consolidated node. 
This reduces the total number of nodes in the global FCM and constitutes a form of ``dimensionality reduction''. 
Every node maps to exactly one consolidated node and no node is discarded.\\
The system prompt for node consolidation instructs the LLM agent to run  global deduplication, thematic clustering, and intra-theme semantic merging.
The agent returns a unified list of nodes for a set of local FCMs \\

\noindent{\bf{Bayesian De-Chunking}:}  The LLM-extracted FCMs from the text chunks mix through convex combination to give the mixed FCM $\mathcal{F}_{\text{Likelihood}}$ as described above.
These likelihood FCMs also give their corresponding posterior FCMs $\mathcal{P}_k$ via equation (\ref{eq:posterior_FCM}). 
The posterior FCMs may not have the same nodes or edges as their corresponding likelihood FCMs.

\section{Simulation Results}

We applied our chunking method to ``The Thucydides Trap'' article  by Graham Allison \cite{allison2015thucydides}.
So $\mathcal{X}$ represents the article (full-text).
This article has 29 paragraphs.

We use the Gemini 3.1 Pro Preview \cite{googledeepmind2026gemini31pro} as the LLM agent for this simulation.
Gemini 3.1 Pro Preview excels at complex problem-solving tasks that require advanced reasoning.
We set the temperature to 0, top $p$ to 1, and top $K$ to 1.0 for all the Gemini API calls. 
We compared the effect of $d_{\max}$ on FCM extractions with our non-overlapping chunking (Algorithm ~\ref{alg:nonoverlapping_chunking}). 
We also compared the effect of $d_{\max}$ and splitting factor $\alpha$ for proportional chunking (Algorithm ~\ref{alg:full-chunking}).

We observed various FCM parameters in our simulations and compared how they respond to text chunking. 
Merged nodes refers to the collection of local node lists from a given set of text chunks and may include repeated nodes. 
Semantic themes are the categories that the thematic clustering step produces. 
Consolidated nodes are the nodes that the intra-theme semantic merging step generates. 
Attractors are the stable long-term states or recurring cycles that an FCM converges to after repeated iterations of its update rule.

We also observed the number of nonzero eigenvalues of the FCM weight matrices. 
All zero eigenvalues imply a degenerate causal structure where the FCM captures no meaningful causal relationships and converges to the trivial null attractor. 
Nonzero eigenvalues indicate that the FCM captures genuine causal relationships and produces richer and more informative dynamics. \\

\noindent{\bf{Text Chunking}:}
We tested our non-overlapping chunking method (Algorithm ~\ref{alg:nonoverlapping_chunking}) and compared it to a baseline that feeds the entire text directly into the LLM agent for FCM extraction.
We considered  $d_{\max} \in \{0,1,2,3,4\}$.
The baseline corresponds to $d_{\max} = 0$.
Table ~\ref{tab:nonoverlap_chunking_tt} compares the baseline FCMs with the ones that the non-overlapping chunking method produces.

LLM agent extraction more information from the Thucydides Trap article with the non-overlapping chunking.
We noticed that the LLM agent captures more information as the number of chunks increases.
The number of chunks equals $2^{d_{\max}}$ so it increases with $d_{\max}$.
 The number of merged nodes, unified nodes, and semantic themes increases with more chunks.

We also repeated the simulation for overlapping chunking.
We tested the proportional overlapping chunks.
We noticed the same trend.
Table ~\ref{tab:overlap_chunking_tt} shows the result for the overlapping text chunking. \\


\noindent{\bf{De-chunking of mixed FCMs}:}
Algorithm~\ref{alg:full-chunking} split the 30-paragraph article into 3 chunks. 
The $1^\text{st}$ chunk contained paragraphs 1-15 and produced the 10-node likelihood FCM $\mathcal{F}_1$. 
The $2^\text{nd}$ chunk contained paragraphs 7-23 and produced the 9-node likelihood FCM $\mathcal{F}_2$. 
The $3^\text{rd}$ chunk contained paragraphs 15-30 and produced the 6-node likelihood FCM $\mathcal{F}_3$. 
The 11-node mixed-FCM $\mathcal{F}_{\text{Likelihood}}$ mixed $\mathcal{F}_1$, $\mathcal{F}_2$, and $\mathcal{F}_3$ with equal convex weights $\frac{1}{3}$. 
$\mathcal{F}_1$, $\mathcal{F}_2$, and $\mathcal{F}_3$ also produced their respective posterior FCMs:  $\mathcal{P}_1$ with 8 nodes, $\mathcal{P}_2$ with 8 nodes, and $\mathcal{P}_3$ with 7 nodes. 
The 3 posterior FCMs then also mixed with equal mixing weights to give the 9-node mixed posterior-FCM $\mathcal{F}_{\text{Posterior}}$. 

We asked the same question to all 8 FCMs:  ``What if the rising power's ambition remains unchecked?'' 
We ask this question by clamping `on' the node ``Rising Power's Ambition \& Entitlement''. 
The FCMs $F_1$ and $F_2$ converged to fixed points while their posterior FCMs $\mathcal{P}_1$ and $\mathcal{P}_2$ converged to limit cycles. 
$\mathcal{P}_1$ converged to a 2-step limit cycle while $\mathcal{P}_2$ converged to a 3-step limit cycle. 
But the ``Likelihood of War'' node stayed on in all 4 equilibria. 
So all 4 FCMs agreed that there would be war.

The FCM $\mathcal{F}_3$ disagreed. 
It converged to a fixed point where the ``Likelihood of War'' node stayed off. 
But its posterior FCM $\mathcal{P}_3$ converged to another fixed point where the ``Likelihood of War'' node stayed on. 
The likelihood-FCM mixture $\mathcal{F}_{\text{Likelihood}}$ converged to the same fixed point as $\mathcal{F}_1$ while the posterior-FCM mixture $\mathcal{F}_{\text{Posterior}}$ converged to the same fixed point as $\mathcal{P}_3$. 
Both FCM-mixtures agreed that there would be war.

The posterior FCM $\mathcal{P}_3$ had 2 edge weights whose magnitude exceeded 1. 
The FCM did not change its equilibria even when those magnitudes were set to 1. 
\vspace{-0.1in}

\begin{table}[t]
\caption{\footnotesize{Thucydides trap and non-overlapping text chunking: FCM extraction with an agentic LLM and non-overlapping text chunking}}\label{tab:nonoverlap_chunking_tt}
\centering 
\begin{tabular}{|l|c|c|c|c|c|}
\hline
 \multirow{1}{*}{Count} & \multirow{1}{*}{$d_{\max} = 0$}  & \multirow{1}{*}{$d_{\max} = 1$}   & \multirow{1}{*}{$d_{\max} = 2$}   & \multirow{1}{*}{$d_{\max} = 3$} & \multirow{1}{*}{$d_{\max} = 4$}  \\
\hline
Chunks  & {1} & {2} & {4} & {8} & {16} \\
Merged nodes & {10 } & {18} & {28} & {52}  & {75} \\
Unified nodes & {9} & {10}  & {19}  & {25}  & {28}\\
Semantic themes & {4} & {4}  & {5} & {5} & {5}\\
Attractors & {1}  & {1}  & {1}  &{6} & {2}\\
Nonzero edges & {10}  & {14} & {23} & {29} & {42}\\
Nonzero eigenvalues & {0} & {0} & {0} & {7} & {7}\\
\hline
\end{tabular}
\vspace{-0.18in}
\end{table}

\begin{table}[t]
\caption{\footnotesize{Thucydides Trap and overlapping text chunking: FCM extraction with an agentic LLM and the overlapping text chunking method (Algorithm \ref{alg:full-chunking}).
We used proportional overlapping chunks with split factor $\alpha = 0.5$. }}\label{tab1}
\label{tab:overlap_chunking_tt}
\centering 
\begin{tabular}{|l|c|c|c|c|}
\hline
\multirow{1}{*}{{Count}} & \multirow{1}{*}{$d_{\max} = 0$}  & \multirow{1}{*}{$d_{\max} = 1$}   & \multirow{1}{*}{$d_{\max} = 2$}   & \multirow{1}{*}{$d_{\max} = 3$} \\
\hline
Chunks   & {1} & {3} & {7} & {15}  \\
Merged nodes & {10 } & {28} & {51} & {96}   \\
Unified nodes & {9} & {11}  & {19}  & {32} \\
Semantic themes & {4} & {5}  & {6} & {6} \\
Attractors & {1}  & {1}  & {2}  &{2} \\
Nonzero edges & {10}  & {16} & {32} & {49} \\
Nonzero eigenvalues & {0} & {0} & {3} & {10} \\
\hline
\end{tabular}
\vspace{-0.2in}
\end{table}

\begin{figure*}[htbp]
\begin{subfigure}{0.5\linewidth}
\centering
\includegraphics[height=0.56\textwidth, width=1.0\textwidth]{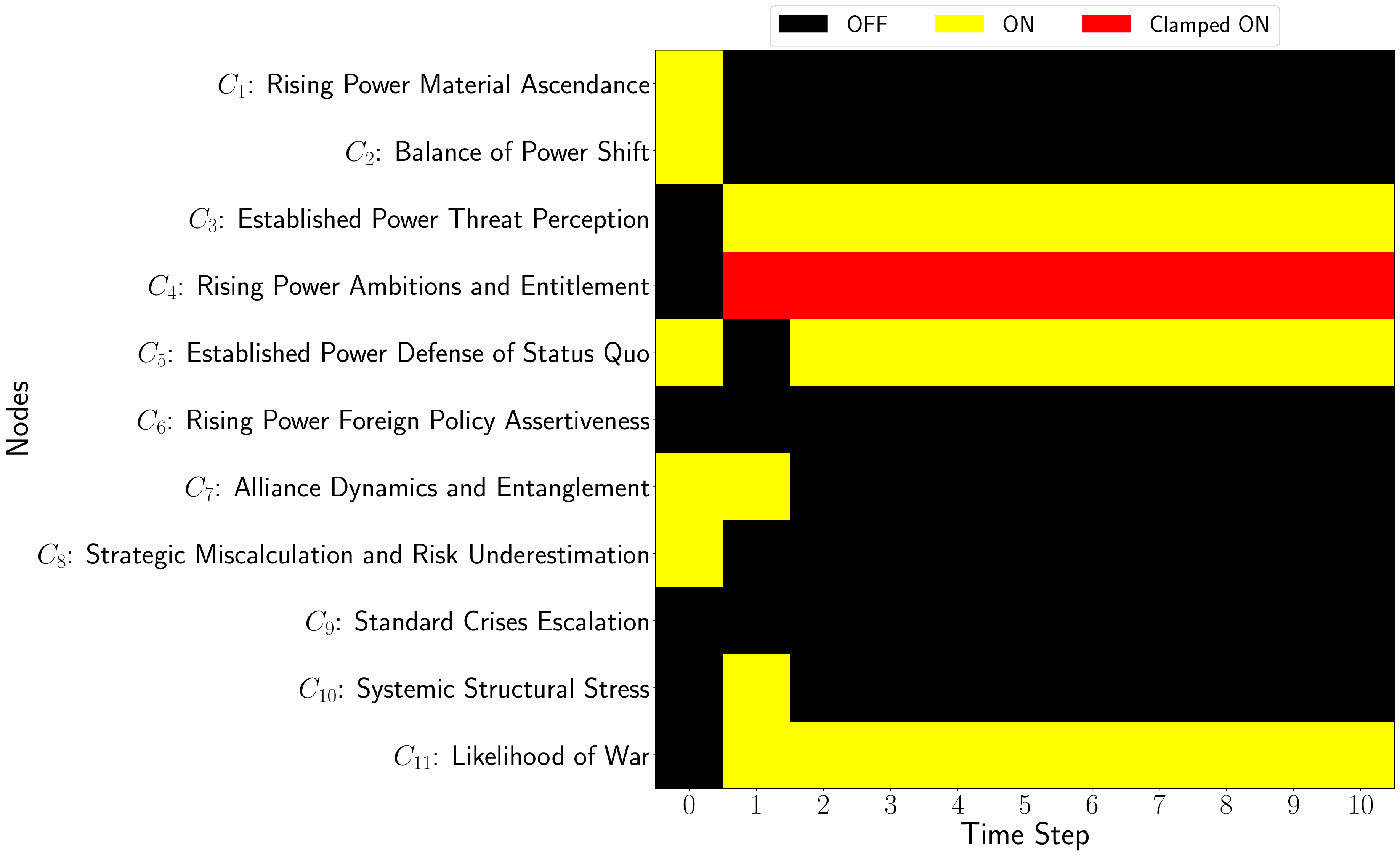}
\caption{{${\mathcal{F}_1}$'s equilibrium}}
\label{fig:F1}
\vspace{0.1in}
\end{subfigure}
\begin{subfigure}{0.5\linewidth}
\centering
\includegraphics[height=0.56\textwidth, width=1.0\textwidth]{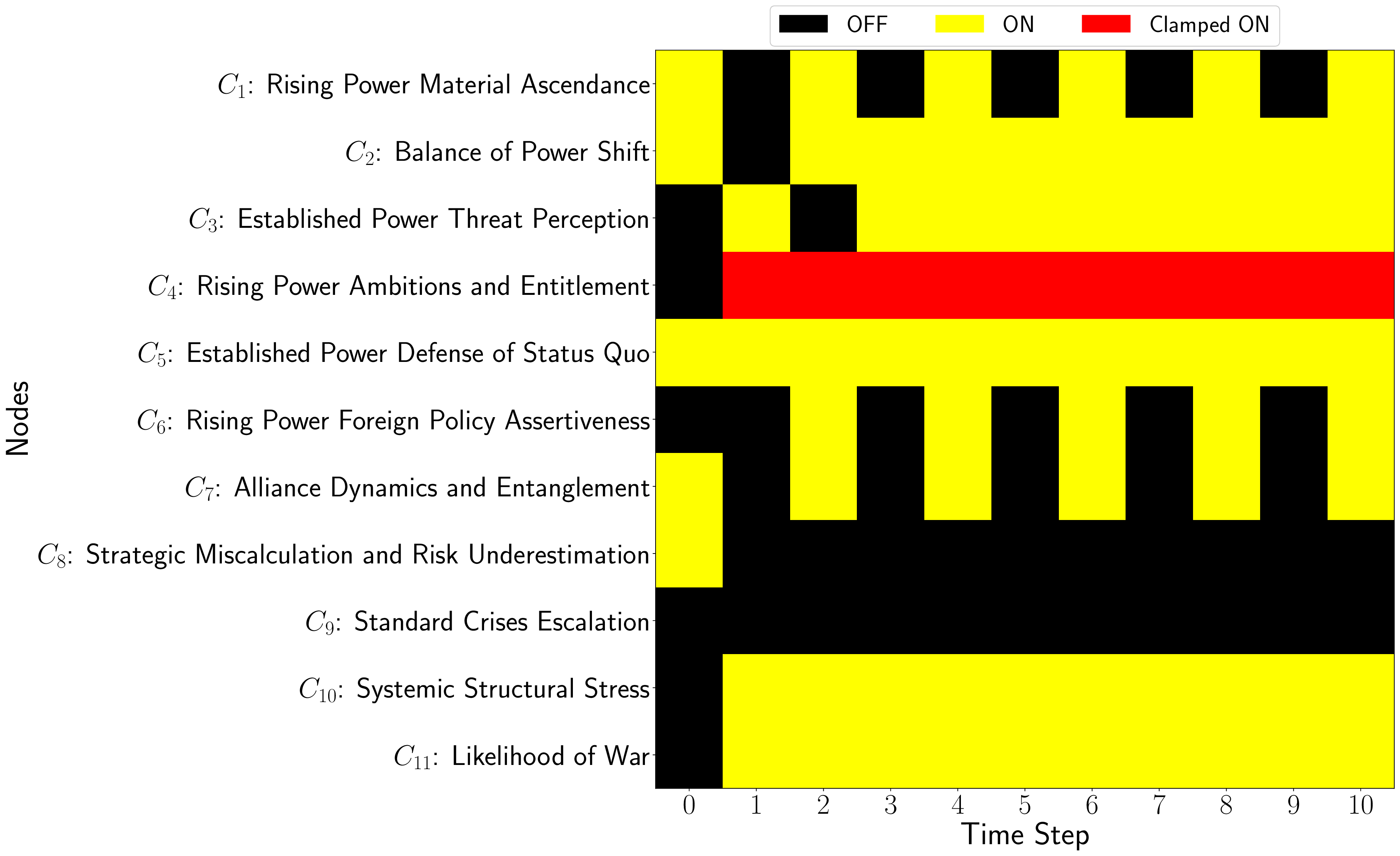}
\caption{{${\mathcal{P}_1}$'s equilibrium}}
\label{fig:P1}
\vspace{0.1in}
\end{subfigure}
\hfill
\begin{subfigure}{0.5\linewidth}
\centering
\includegraphics[height=0.56\textwidth, width=1.0\textwidth]{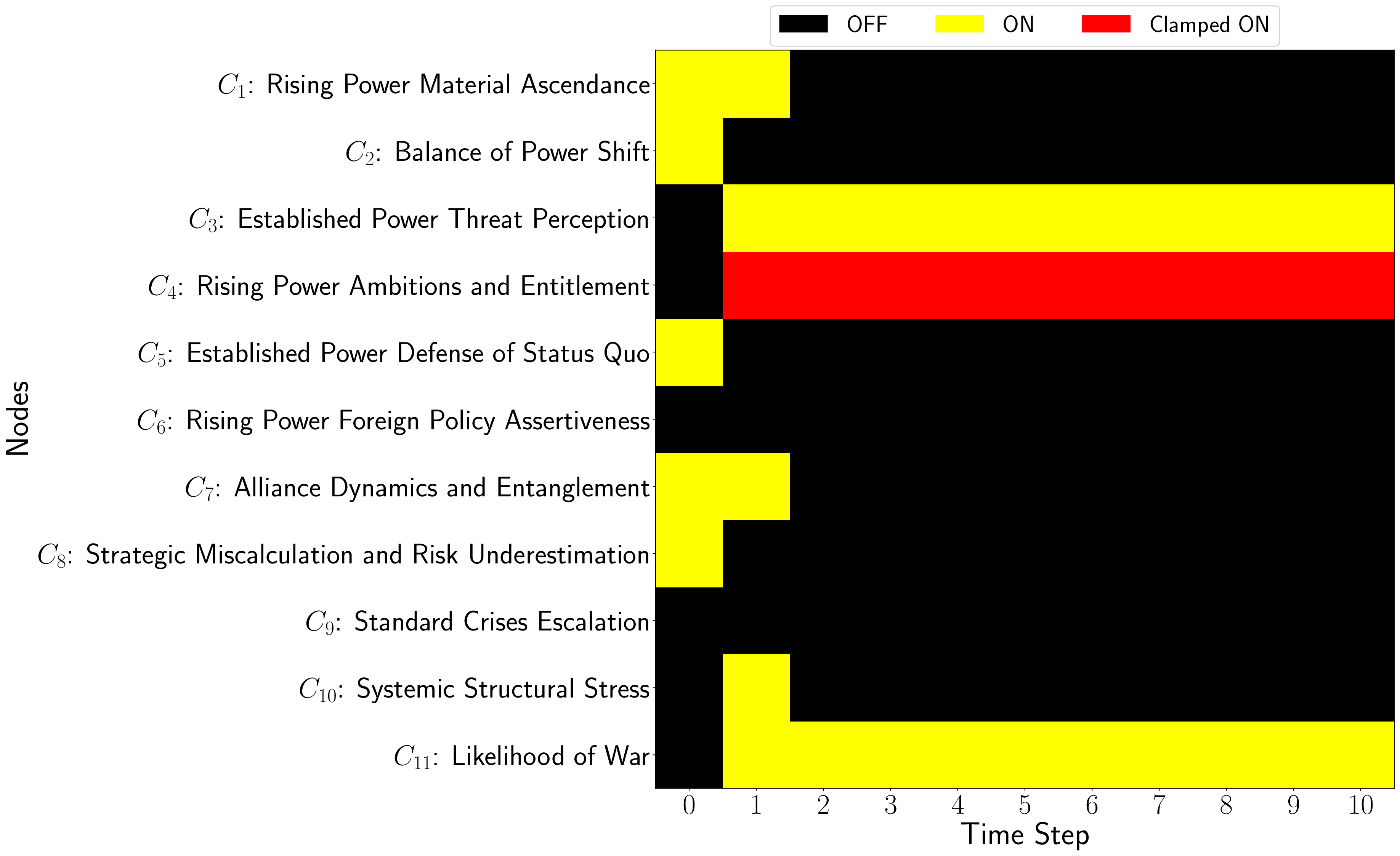}
\caption{{${\mathcal{F}_2}$'s equilibrium} }
\label{fig:F2}
\vspace{0.1in}
\end{subfigure}
\hfill
\begin{subfigure}{0.5\linewidth}
\centering
\includegraphics[height=0.56\textwidth, width=1.0\textwidth]{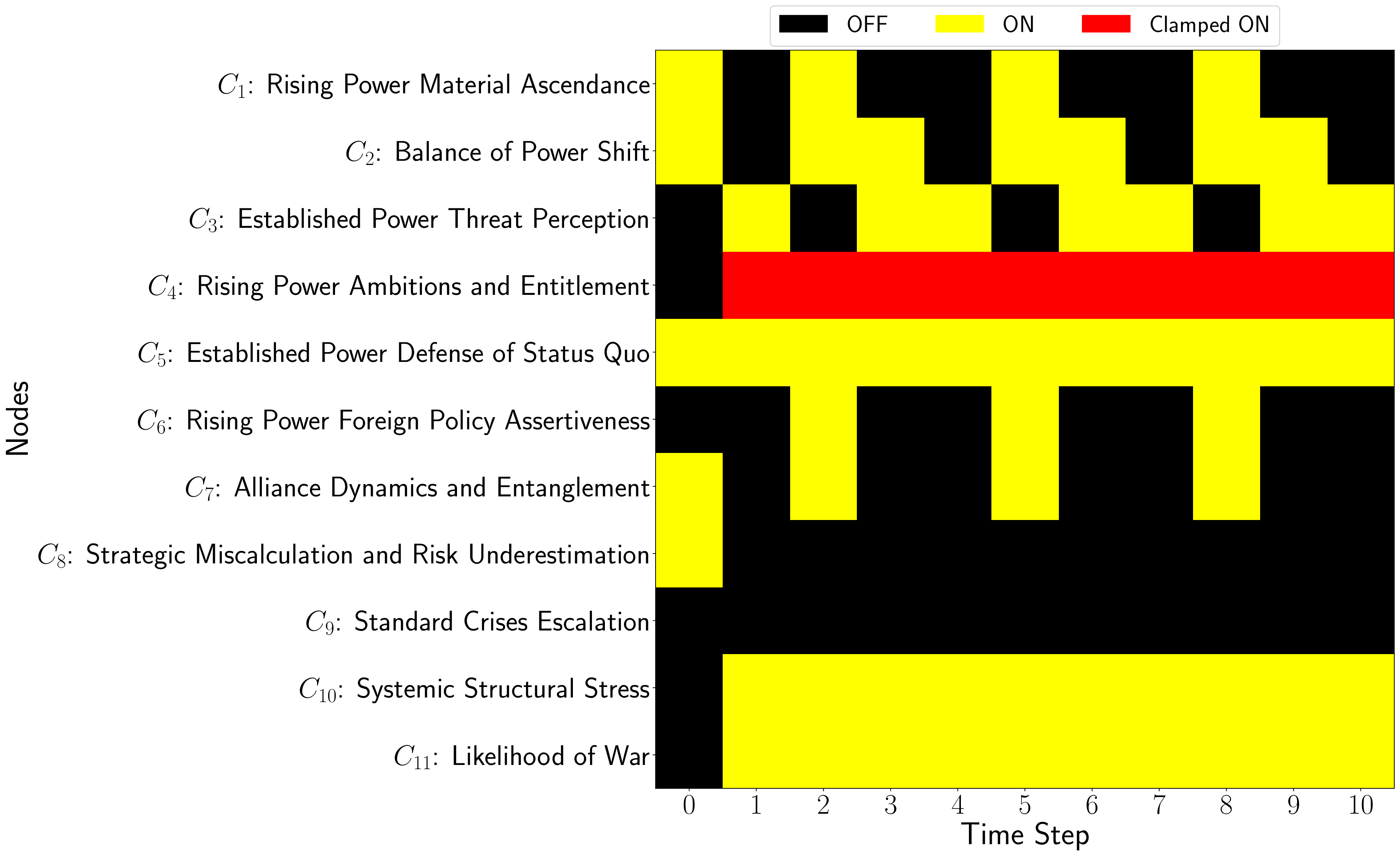}
\caption{{${\mathcal{P}_2}$'s equilibrium} }
\label{fig:P2}
\vspace{0.1in}
\end{subfigure}
\begin{subfigure}{0.5\linewidth}
\includegraphics[height=0.56\textwidth, width=1.0\textwidth]{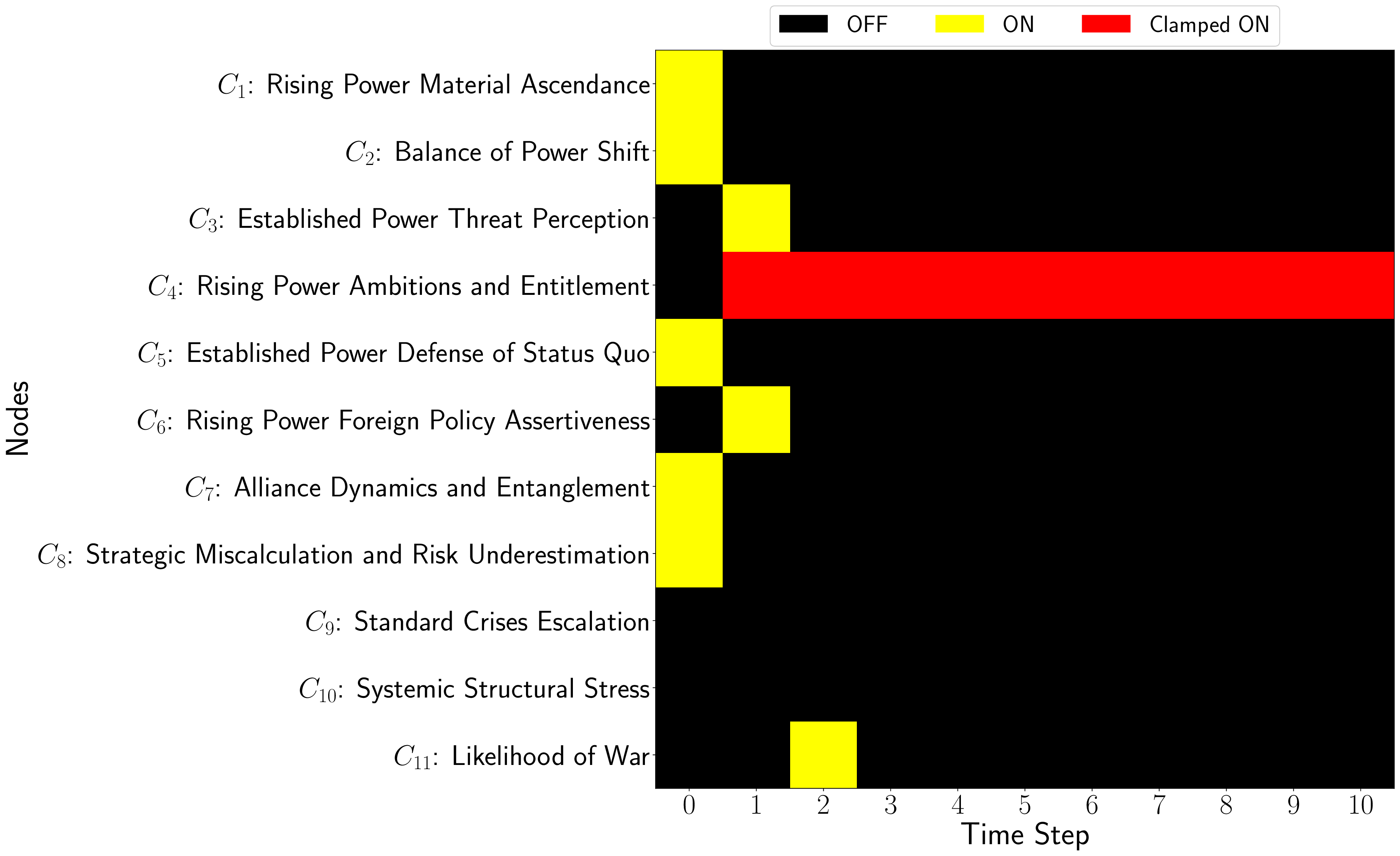}
\caption{{${\mathcal{F}_3}$'s equilibrium}}
\label{fig:F3}
\vspace{0.1in}
\end{subfigure}
\hfill
\begin{subfigure}{0.5\linewidth}
\centering
\includegraphics[height=0.56\textwidth, width=1.0\textwidth]{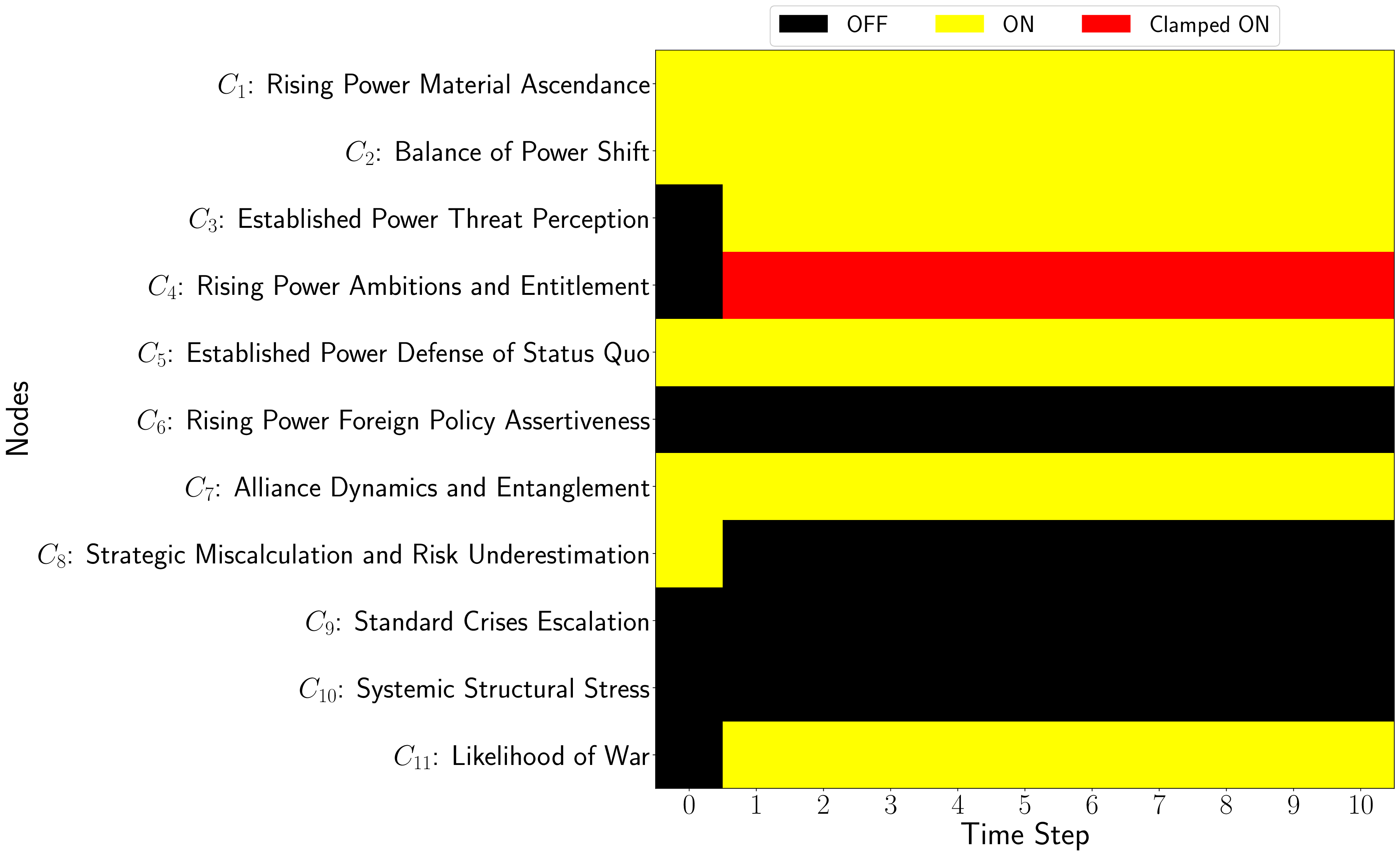}
\caption{{${\mathcal{P}_3}$'s equilibrium}}
\label{fig:P3}
\vspace{0.1in}
\end{subfigure}
\hfill
\begin{subfigure}{0.5\linewidth}
\centering
\includegraphics[height=0.56\textwidth, width=1.0\textwidth]{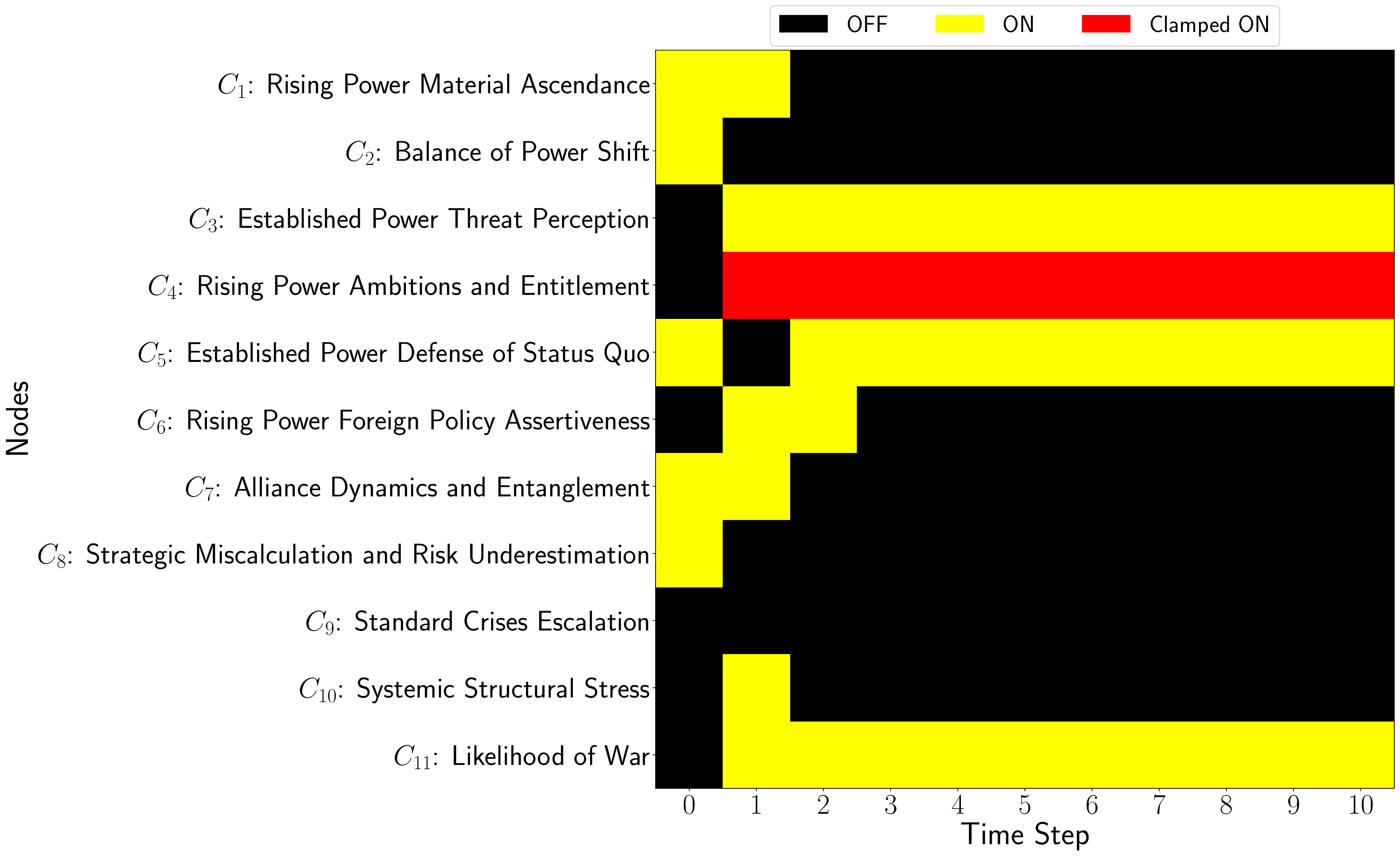}
\caption{{$\mathcal{F}_{\text{Likelihood}}$'s equilibrium}}
\label{fig:F}
\vspace{0.1in}
\end{subfigure}
\hfill
\begin{subfigure}{0.5\linewidth}
\centering
\includegraphics[height=0.56\textwidth, width=1.0\textwidth]{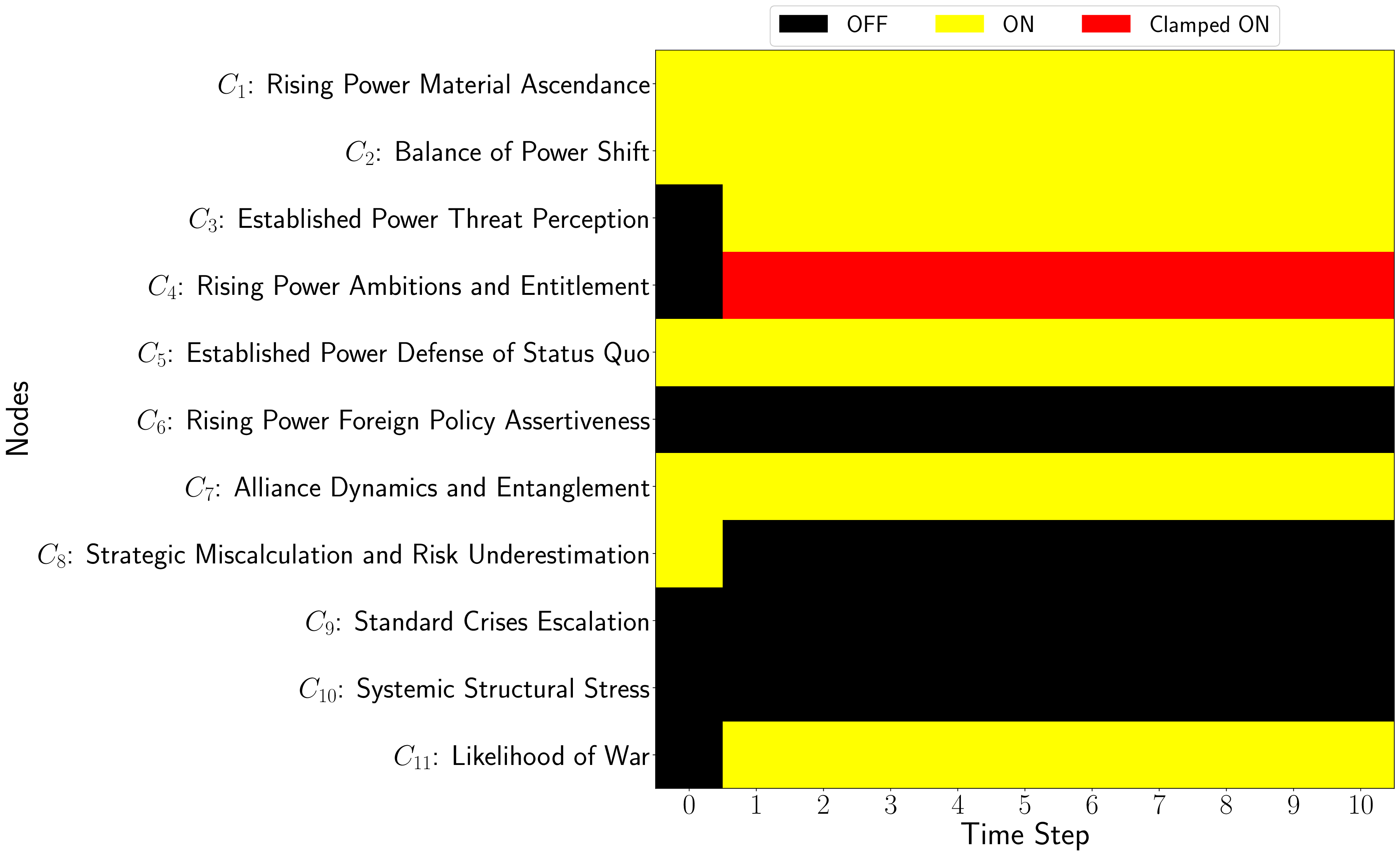}
\caption{{$\mathcal{F}_{\text{Posterior}}$'s equilibrium}}
\label{fig:P}
\vspace{0.1in}
\end{subfigure}
\caption{\footnotesize{Equilibrium answers to the clamping question:  What if the rising power's ambitions remain unchecked? 
The time steps are along the $x$-axis and the nodes are along the $y$-axis. 
The purple nodes are inactive and the yellow nodes are active. 
The clamped-on nodes are in red. 
(a), (c), and (e) show the equilibria from the 3 likelihood FCMs $\mathcal{F}_1$, $\mathcal{F}_2$, and $\mathcal{F}_3$ and (g) shows the equilibrium from their equal weight mixture $\mathcal{F}_{\text{Likelihood}}$. 
(b), (d), and (f) show the equilibria from the 3 posterior FCMs $\mathcal{P}_1$, $\mathcal{P}_2$, and $\mathcal{P}_3$ and (h) shows the equilibrium from their mixture $\mathcal{F}_{\text{Posterior}}$. 
The last node ``Likelihood of War'' stays on in every equilibrium except (e). }}
\end{figure*}

\section{Conclusion}

Controlled intelligent agents can map text to text chunks and then map the chunks to sparse causal edge matrices that define feedback fuzzy cognitive maps.
The chunk matrices naturally mix or combine to produce a likelihood-like FCM that represents the feedback causal structure of the entire sampled document. 
The mixing weights must be convex and can depend on input values or otherwise vary with time.
The mixing structure naturally gives an operator-level set of Bayesian posterior FCM matrices (which may require clipping or normalizing some edge values).
These document FCMs can then mix to represent larger documents.
They can also combine in hierarchies to store and fire documents in a structure large knowledge base.

%
%
%
%
\bibliographystyle{splncs04}
\bibliography{bibdata}
\end{document}